\documentclass[letterpaper,twocolumn]{article}

\usepackage{times}
\usepackage{helvet}
\usepackage{courier}

\usepackage{color}
\usepackage{graphicx}
\usepackage{booktabs}
\usepackage[vlined,ruled]{algorithm2e} 

\newcommand\mycommfont[1]{\normalfont{#1}}

\begin{document}
\title{Batched Lazy Decision Trees}

\author{Mathieu Guillame-Bert and Artur Dubrawski\\ \\
       Carnegie Mellon University\\
        School of Computer Science\\
        Pittsburgh, United States\\
        mathieug@andrew.cmu.edu, awd@cs.cmu.edu
        }

\maketitle
\begin{abstract}
\begin{quote}
We introduce a batched lazy algorithm for supervised classification using decision trees. It avoids unnecessary visits to irrelevant nodes when it is used to make predictions with either eagerly or lazily trained decision trees. A set of experiments demonstrate that the proposed algorithm can outperform both the conventional and lazy decision tree algorithms in terms of computation time as well as memory consumption, without compromising accuracy. 
\end{quote}
\end{abstract}

\section{Introduction}
In machine learning, the task of a classification algorithm is to predict the class label of an unlabeled observation, given an available collection of labeled training observations. Eager classification algorithms solve this task by building an intermediate explicit predictive model that can be then used to adjudicate the unlabeled observations. Conversely, non-eager classification algorithms (also called lazy classification algorithms) do not build an intermediate model. Instead, they compare the unlabeled observations to the whole set of training observations when making a prediction. As an example, ID3 decision tree and Random Forest~\cite{Breiman2001} are eager algorithms, while k-Nearest Neighbours is an example of a lazy algorithm.

Popular decision tree algorithms and more recently developed boosted and bagged decision tree algorithms are examples of eager algorithms that have been shown to be powerful state of the art methods. These algorithms build one or several decision trees, and use them as predictive models to process future queries. Such models may be computationally expensive to build but they are cheap to apply when trained.

Friedman et al.~\cite{Friedman1996} and later Fern and Brodley~\cite{Fern2003} experimented with the emulation of decision tree algorithms in lazy frameworks. These two approaches are respectively called a lazy decision tree and a boosted lazy decision tree. Unlike conventional (eager) decision tree algorithms, these approaches do not build full decision trees from training data but only explore a single path of the decision tree per unlabeled observation. While lazy decision trees can perfectly emulate operations of the conventional decision trees, Friedman et al.~\cite{Friedman1996} and Fern and Brodley~\cite{Fern2003} have shown how using mutual information between training set and the unlabeled data during the decision tree building can improve the algorithm's accuracy.

The main drawback of lazy decision trees is their high computational cost in the prediction phase. They need to be re-run independently for each unlabeled observation. A potential solution to that problem is mentioned by Friedman et al. and it considers introducing of a caching scheme'. With this scheme, the algorithm remembers and re-uses the exploration paths traversed in previous runs. No details are available about this scheme. The authors only report that "it consumes a lot of memory''~\cite{Friedman1996}.

In this paper, we introduce a lazy decision tree algorithm that exactly emulates the logic of eager decision trees and that solves the problem of unnecessary node exploration of both the conventional and lazy decision tree algorithms. The proposed algorithm is faster than eager decision trees, and therefore faster than lazy decision trees, while consuming the same amount of memory as a lazy decision tree (i.e.\ only one path through the tree is kept in memory at any instant) without the need for using memory-consuming caching schemes. We will refer to this algorithm as a batched lazy decision tree. Performance of the proposed algorithm is evaluated using bagging the models into ensembles of decision trees (bootstrap aggregation).

\section{Decision trees and lazy decision trees}
\label{sec:dt_and_lazy_dt}

(Bagged) eager decision trees work by growing a set of non-pruned decision trees--each one trained on a bootstrapped set of reference observations. Listing~\ref{algo:dt} shows a generic bagged decision tree algorithm. A lazy decision tree emulates that operation by independently growing a single tree branch to handle each unlabeled observation. Listing~\ref{algo:ldt} shows a generic bagged lazy decision tree algorithm.

\begin{figure}
\centering
\includegraphics[scale=0.92]{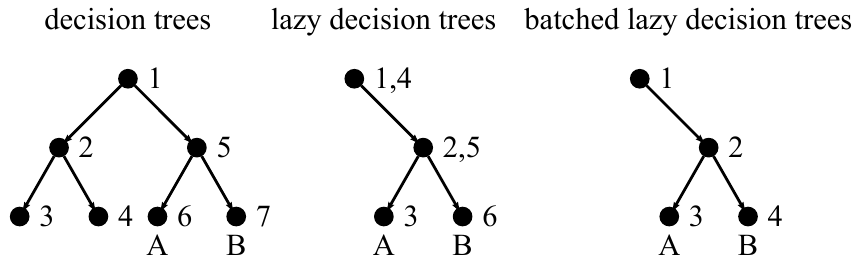}
\caption{Example of exploration of the same decision tree depending on the algorithm used. The example considers a training set without cached representation and two unlabeled observations A and B. The indices to the right of each node represent the sequences of exploration of the nodes (i.e.\ the step index of an algorithm).}
\label{fig:exploation}
\end{figure}

Since lazy decision trees only explore the branches used to predict unlabeled observations, they may end up exploring fewer nodes than exist in the equivalent conventional eager decision trees. This may happen if the test set is approximately just a subset of the data used for training the model. Nonetheless, lazy decision trees will repeat the exploration of some nodes. In practice, unless the number of unlabeled observations is small, lazy decision trees can be slower in prediction mode than the equivalent eager decision trees. Both algorithms are illustrated in Figure~\ref{fig:exploation}.

We will now focus on the time and memory complexity of these algorithms. We consider a training dataset with $n$ observations and $m$ attributes. We assume the time complexity of evaluating an attribute is linear in the number of observations. Therefore, using a $k$-fold cross-validation scheme and expanding nodes with a minimum of $p$ training observations in them, the time complexity of an eager decision tree model is $O( \sum_{i=0}^{\log_2 \frac{n}{p} } 2^i m \frac{n}{2^i} ) = O(k m n \log_2 \frac{n}{p} )$ and the time complexity of a lazy decision tree model is $O( \sum_{i=0}^{\log_2 \frac{n}{p} } m \frac{n}{2^i} ) = O(m n^2)$ on average. Note that $\log_2 \frac{n}{p}$ is the average depth of exploration, $2^i$ is the average number of nodes at depth $i$, and $\frac{n}{2^i}$ is the average number of training set observations contained in a node at depth $i$.

Memory is rarely an issue when implementing decision trees but it deserves characterization. Building an eager decision tree requires memory for the stack operations used during the recursive construction of the tree as well as some storage for the final trained model. Lazy decision trees only require the stack. Using the same notation, the stack size and model size of decision trees are respectively $O( \sum_{i=0}^{\log_2 \frac{n}{p} } \frac{n}{2^i} ) = O(n)$ and $O(\frac{n}{p})$ (i.e.\ number of non-leaf nodes in a decision tree) on average. Note that increasing the number of bootstrap samples in a bagged decision tree does not impact the stack size but just linearly increases the memory required to store the resulting trees. The stack size of lazy decision trees is $O(n)$.

Note that our analytic results assume that $\log_2 \frac{n}{p} \leq d$ where $d$ is the maximum depth of exploration.

\begin{algorithm}
\DontPrintSemicolon
\caption{Generic decision tree algorithm with bagging}
\SetKwInOut{Input}{input}\SetKwInOut{Output}{output}
\Input{
$T \in M^{n \times (m+1)}$ : The training set with $n$ observations and $m$ attributes. The last column contains the class labels.\\
$b$ : Number of bootstraps.\\
$c$ : Minimum number of observations in a node (e.g. 5).\\
$d$ : Maximum depth (e.g. 20).\\
}
\Output{
$R$ : A set of bagged decision trees.
}

\SetKwFunction{proc}{rec\_build}

\SetKwBlock{myalg}{Algorithm}{}
\SetKwBlock{myproc}{Subroutine \proc{$T' \in M^{n' \times (m+1)}$,$i$}
}{}

\BlankLine
\myalg
{
$R \leftarrow \emptyset$\;
\For(\tcp*[f]{\mycommfont For each bootstrap}){$ i\ \leftarrow 1$ \KwTo $b$}{
	$T' \leftarrow $ bootstrap rows of $T$\;
    $R \leftarrow R \cup \{ \proc(T',0) \}$\;
	}
}

\myproc
{
	\If{$i >d$ or $n'<c$ or $T'$ is pure}{
		$t \leftarrow $ \textbf{create} a node labeled with the most frequent class in $T'$\;
        \KwRet $t$\;
        }
    $a \leftarrow$ \textbf{find} the condition of $T'$ with the highest information gain\;
    $L \leftarrow $ subset of $T'$ rows' such that $a$ is invalid\;
    $R \leftarrow $ subset of $T'$ rows' such that $a$ is valid\;
	$t_l \leftarrow \proc{L,i+1}$\;
	$t_r \leftarrow \proc{R,i+1}$\;
    $t \leftarrow $ \textbf{create} a node with two children $t_l$ and $t_r$, and labeled with $a$.\;
}

\label{algo:dt}
\end{algorithm}

\begin{algorithm}
\DontPrintSemicolon
\caption{Generic lazy decision tree algorithm with bagging}
\SetKwInOut{Input}{input}\SetKwInOut{Output}{output}
\Input{
$T \in M^{n_t \times (m+1)}$ : The training set with $n$ observations and $m$ attributes. The last column contains the class labels.\\
$S \in M^{n_s \times m}$ : The testing set.\\
$b$ : Number of bootstraps.\\
$c$ : Minimum number of observations in a node (e.g. 5).\\
$d$ : Maximum depth (e.g. 20).\\
}
\Output{
$R \in M^{n_s \times h}$ : The predicted probability for each class and each row of the testing dataset.
}
\SetKwBlock{myalg}{Algorithm}{}{}

\BlankLine
\myalg
{

\For(\tcp*[f]{\mycommfont For each bootstrap}){$ i\ \leftarrow 1$ \KwTo $b$}
	{
	$T' \leftarrow $ bootstrap rows of $T$\;
	\For(\tcp*[f]{\mycommfont For each test observation}){$ j\ \leftarrow 1$ \KwTo $n_s$ }
		{
		$X \leftarrow T'$\;
		$d' \leftarrow 0$\;
		\While{$nrow(X) \geq c$ and $d' \leq d$ and $X$ is not pure}
			{
			$a \leftarrow$ \textbf{find} the condition of $X$ with the highest information gain\;
			$d' \leftarrow d' + 1$ \;
			\If{condition $a$ is valid for the test observation $j$}
				{
				$X \leftarrow $ subset of $X$ rows such that $a$ is valid\;
				}
			\Else
				{
				$X \leftarrow $ subset of $X$ rows such that $a$ is invalid\;
				}
			}	
		$f \leftarrow $ the most represented class in $X$\;			
		$R_{i,f} \leftarrow R_{i,f} + 1 / b$ \;
		}
	}
}
\label{algo:ldt}
\end{algorithm}

\section{Batched lazy decision trees}
\label{sec:batched_lazy_dt}

We will now introduce our batched lazy decision tree algorithm. We begin with the listing, and then discuss and compare batched lazy decision trees with the eager and lazy decision trees.

Listing~\ref{algo:bldt} shows the batched lazy decision tree algorithm with bagging. Batched lazy decision trees behave similarly to lazy decision trees, however a lazy decision tree explores one path of inference going through it at a time. Since different test observations are likely to end up in different leaf nodes, lazy decision tree algorithm needs to be re-executed for each test observation. Batched lazy decision tree algorithm solves this issue by enabling the exploration of an arbitrary sub-tree instead of a single path. Batched lazy decision tree explores the same nodes as the regular lazy decision trees, however, each necessary node is visited only once through a single pass. Batched lazy decision trees will not visit nodes that are not used for the predictions. A diagram representation of a batched lazy decision tree is shown in Figure~\ref{fig:exploation}.
Note that the test data is never sorted in any way for the batched lazy algorithm. We only require the property that a test observation is being assigned to one and only one child of a given non-leaf node. 

The time complexity of batched lazy decision trees is $O(\sum_{i=0}^{\frac ip} 2^i ( 1 - ( 1 - 2^{-i} )^{\frac{n}{k}} ) ( \frac{n-\frac{n}{k}}{2^i} + \frac{\frac{n}{k}}{i^2} ) = O(m n ( \log_ 2 \frac{n}{p} - \sum_{i=0}^{\log_2{\frac{n}{p}}} (1-2^{-i})^\frac{n}{k} ) )$. The computational complexity is difficult to characterize analytically, but it is strictly lower than the computational complexity of eager decision trees. The memory complexity is $O(n)$, similar to that of lazy decision trees.

\begin{algorithm}
\DontPrintSemicolon
\caption{Batched Lazy Decision Tree algorithm with bagging}
\SetKwInOut{Input}{input}\SetKwInOut{Output}{output}
\Input{
$T \in M^{n_t \times (m+1)}$ : The training set with $n$ observations and $m$ attributes. The last column contains the class labels.\\
$S \in M^{n_s \times m}$ : The testing set.\\
$b$ : Number of bootstraps.\\
$c$ : Minimum number of observations in a node (e.g. 5).\\
$d$ : Maximum depth (e.g. 20).\\
}
\Output{
$R \in M^{n_s \times h}$ : The predicted probability for each class and each row of the testing dataset.
}

\SetKwFunction{proc}{rec\_run}

\SetKwBlock{myalg}{Algorithm}{}{}

\SetKwBlock{myproc}{Subroutine  \proc{
  $T' \in M^{n'_t \times (m+1)}$
, $S' \in M^{n'_s \times m}$
, $d'$
} }{}{}

\myalg{
\For(\tcp*[f]{\mycommfont For each bootstrap}){$ i\ \leftarrow 1$ \KwTo $b$}
	{
	$T' \leftarrow $ bootstrap rows of $T$\;
	$\proc(T',S,0)$\;
	}
}

\myproc
{

	\If{$d' >d$ or $n'_t<c$ or $T'$ is pure}{
		$f \leftarrow $ the most frequent class in $T'$\;	
		
		\For(\tcp*[f]{\mycommfont For each test observation}){$ j\ \leftarrow 1$ \KwTo $n'_s$}
			 {
			 $R_{j,f} \leftarrow R_{j,f} + 1 / b$ \;
			 }
		
        \KwRet \;
        }
        
    $a \leftarrow$ \textbf{find} the condition of $T'$ with the highest information gain\;

	$L_s \leftarrow $ subset of $S'$ rows such that $a$ is invalid\;
    $R_s \leftarrow $ subset of $S'$ rows such that $a$ is valid\;
    
    \If{$L_s$ is not empty}
    	{
    	$L_t \leftarrow $ the subset of $T'$ rows such the condition $a$ is invalid\;
    	\proc{$L_t$,$L_s$,$d'+1$}\;
    	}
    	
    \If{$R_s$ is not empty}
    	{
    	$R_t \leftarrow $ the subset of $T'$ rows such the condition $a$ is valid\;
    	\proc{$R_t$,$R_s$,$d'+1$}\;
    	}
 }
\label{algo:bldt}
\end{algorithm}

\section{Empirical comparison}
\label{sec:experiments}

In this this section, we empirically evaluate the computational costs of eager, regular lazy, and batched lazy algorithms on five publicly available datasets--each dataset having different characteristics. Each algorithm is configured to compute $b=100$ bootstraps, explore nodes with the minimum number of training data observations of $p=5$ and a maximum depth of exploration of $d=20$. Note that the computation complexities of all algorithms are linearly dependent to the parameter $b$. Also, both $p$ and $d$ limit the exploration capability of the tree. If $\log_2 \frac np < d$, it is the $p$ parameter that will mainly limit the exploratory behaviour. This constrain is true in all the tested datasets.

\subsection{Datasets and implementation}

We consider a diverse set of five benchmark datasets. Three popular UCI datasets~\cite{uci} (Breast, Adult, and Gamma) and two medical genetic datasets (C2U~\cite{ctou} and ALL~\cite{all}). Table~\ref{tab:datasets} reports various statistics for these sets. Informally, Breast, Adult, Gamma and C2U are small and average sized common type datasets, while ALL is a typical genetic dataset that contains many fewer observations than attributes.

\begin{table}
\centering
\caption{Datasets used for the experimental evaluation.}
\label{tab:datasets}
\begin{tabular}{@{}lrrrr@{}}
\toprule
Dataset & Rows & Columns & Rows/Cols & Size (kB) \\ \midrule
C2U & 2695 & 212 & 12.7 & 1139 \\
Breast & 569 & 31 & 18.4 & 118 \\
Adult & 32561 & 15 & 2 170 & 1194 \\
All & 129 & 12626 & 0.01 & 26780 \\
Gamma & 19020 & 11 & 1 729 & 1443 \\ \bottomrule
\end{tabular}
\end{table}

All three algorithms have been implemented in C++ and run on the same Intel-I7 computer equipped with 16GB of main memory. For accuracy of the results, the reported computation times are the sum of the user and kernel CPU times. All three algorithms rely on the same function to determine the best split of data at each node according to the information gain provided by each attribute.

\subsection{Results}

\begin{figure}
\centering
\includegraphics[width=\columnwidth]{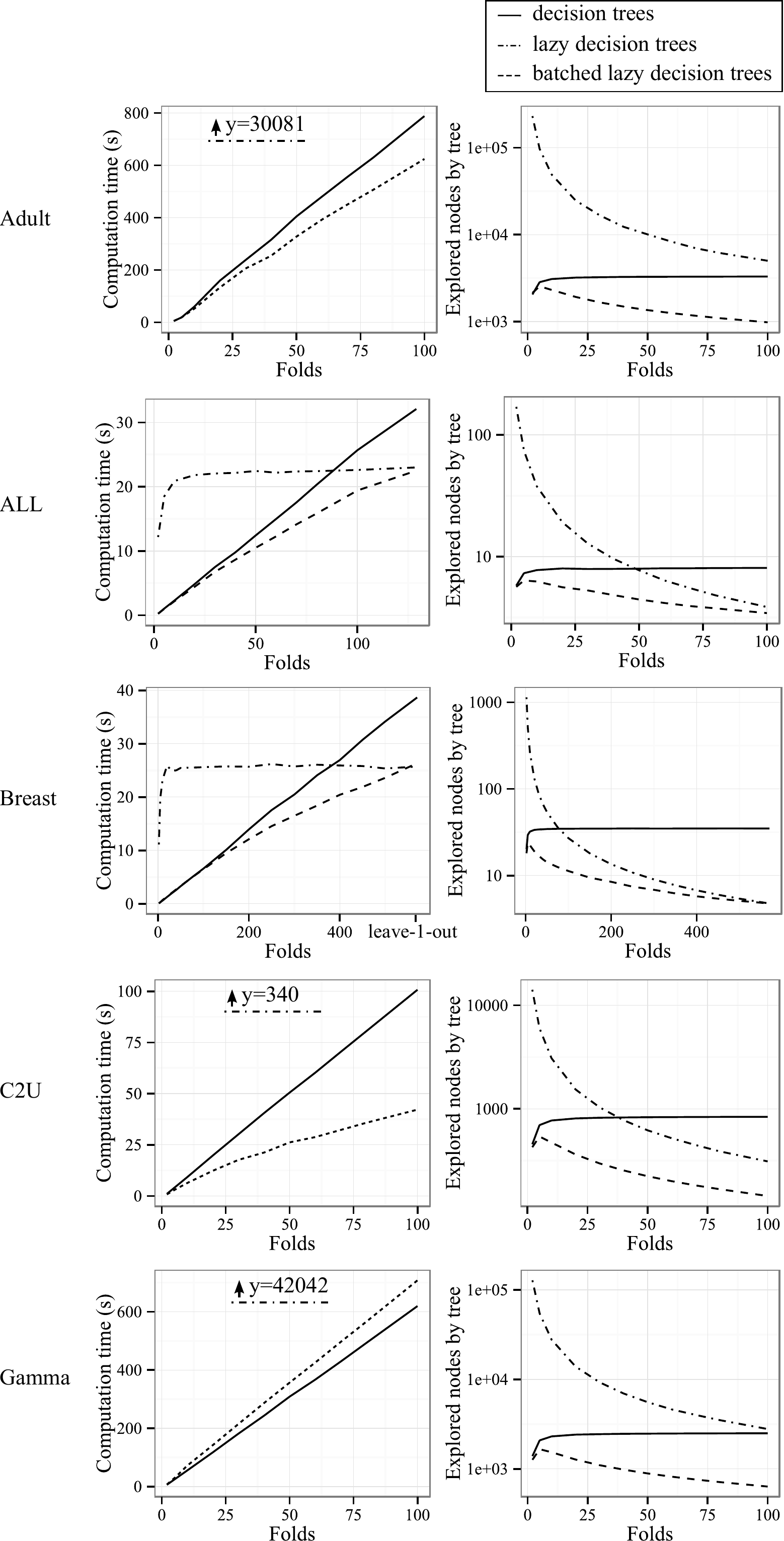}
\caption{Computation time and average number of explored nodes as a function of the number of cross-validation folds.}
\label{fig:results}
\end{figure}

Figure~\ref{fig:results} shows the computation times of k-fold cross-validation for each algorithm and for various values of k, for each dataset. 
It also shows the average number of nodes explored by each of the considered algorithms.

Table~\ref{tab:results} shows the computation time that each method requires to process one fold in a leave-one-out cross-validation scheme, i.e.\ the time needed to train the model and predict for a single test observation. Finally, table~\ref{tab:results} reports the numerical values from Figure~\ref{fig:results} for 10-fold and 40-fold cross-validation protocols.
Table~\ref{tab:memory} shows the counts of memory words required to train and store (in the case of eager decision trees) one model built during a 10-fold cross-validation scheme. This table supposes that storing one observation index requires one word, and that storing one decision tree node requires four words (i.e.\ the attribute used, condition parameter, and addresses of the left and right sub-nodes).

\begin{table*}
\centering
\caption{Computation time for one test observation (equivalent to the leave-one-out cross-validation protocol), 10-fold, and 40-fold cross-validation for the eager decision tree (DT), the lazy decision tree (L-DT), and the batched lazy decision tree (BL-DT).}
\label{tab:results}
\begin{tabular}{@{}lrrrrrrrrrrr@{}}
\toprule
Dataset & Rows & Columns & \multicolumn{3}{r}{CPU time of one observation} & \multicolumn{3}{r}{CPU time of 10-folds CV} & \multicolumn{3}{r}{CPU time of 40-folds CV} \\ \cmidrule(l){4-12} 
 &  &  & DT & L/BL-DT & Ratio & DT & L-DT & BL-DT & DT & L-DT & BL-DT \\ \midrule
C2U & 2695 & 212 & 1.008 & 0.126 & 8.0 & 9.25 & 340.82 & 6.37 & 40.50 & 340.82 & 21.22 \\
Breast & 596 & 31 & 0.066 & 0.045 & 1.5 & 0.58 & 23.20 & 0.69 & 2.57 & 26.62 & 2.68 \\
Adult & 32561 & 15 & 4.708 & 0.924 & 5.1 & 59.64 & 30 081 & 52.06 & 314.97 & 30081 & 254.02 \\
All & 129 & 12626 & 0.245 & 0.172 & 1.4 & 2.28 & 20.89 & 2.17 & 9.78 & 22.15 & 8.64 \\
Gamma & 19020 & 11 & 6.105 & 2.210 & 2.8 & 55.46 & 42 042 & 71.99 & 243.30 & 42042 & 286.73 \\ \bottomrule
\end{tabular}
\end{table*}

\begin{table}
\centering
\caption{Number of memory words required to train one model in a 10-fold cross validation scheme for the eager decision tree (DT), the lazy decision tree (L-DT), and the batched lazy decision tree (BL-DT).}
\label{tab:memory}
\begin{tabular}{@{}lrrrr@{}}
\toprule
Dataset & \multicolumn{3}{c}{Stack} & Model \\ \cmidrule(l){2-4} \cmidrule(l){5-5}
 & DT & L-DT & BL-DT & DT \\
 \midrule
C2U & 15489 & 4900 & 11683 & 312217 \\
Breast & 2562 & 1036 & 1934 & 13000 \\
Adult & 214435 & 89620 & 170549 & 1243262 \\
All & 499 & 234 & 372 & 3076 \\
Gamma & 148078 & 34582 & 139098 & 927200 \\ \bottomrule
\end{tabular}
\end{table}

First, as Friedman et al.\ have suggested~\cite{Friedman1996}, we observed that lazy decision trees can be impractically slow. More precisely, lazy decision trees tend to require time during cross-validation when the number of folds is small in comparison to the number of observations. E.g.\ for the Breast dataset, lazy decision tree only becomes faster than an eager decision tree when the number folds exceeds 390. On this same dataset, we see that the lazy decision tree and our batched lazy decision tree are computationally equivalent for leave-one-out cross-validation. Additionally, both algorithms are faster than eager decision trees in this same setup. These last two observations are true on all datasets.

We observe that batched lazy decision tree model is strictly faster (Adult, All, C2U), equivalent (Breast), but it can also be slower (Gamma) than regular decision trees for less than a 100-fold cross-validation. In the Gamma dataset, we observe batched lazy decision tree becoming faster than alternatives for the number of folds higher than 350 (not shown in Figure~\ref{fig:results}).

We see that the number of nodes explored by batched lazy decision tree is significantly lower that the number of nodes explored by eager or lazy decision trees. Interestingly, the drop in the number of explored nodes of a batched lazy decision tree in comparison to the eager decision trees is much greater than the computational time improvement. This phenomenon is due to the difference of computation cost of each node: Because of the variability of the number of available observations, nodes at the top of the tree (i.e.\ near the root) are more expensive to compute than nodes at the bottom of the tree (i.e.\ near the leafs). And, most of the node filtering of batched lazy decision trees happens at the nodes at the bottom of the trees.

Finally, we observe that eager decision trees require significantly more stack memory than lazy decision trees, and at a somewhat smaller scale, also more than batched decision trees. Interestingly, we see that storing a bagged decision tree model requires significantly more memory than the stack space required for training the models.

\section{Discussion}
\label{sec:conclusion}
We introduced a lazy algorithm for decision trees. It solves the problem of unnecessary node exploration of both the eager and lazy decision tree algorithms.

W have conducted a set of experiments to evaluate empirically the time complexity of the proposed algorithm and its two contenders. These experiments have shown that the proposed algorithm is faster than both the conventional (eager, not-lazy) decision tree algorithms, and lazy decision tree algorithms -- independently of the number of folds of data and the number of test observations. Finally, we characterized analytically and empirically the memory requirement of each algorithm. And we demonstrated that batched decision trees require significantly less memory that decision trees trained in standard ways. 

\bibliographystyle{plain}
\bibliography{manuscript}

\end{document}